# Automated Vehicle's behavior decision making using deep reinforcement learning and high-fidelity simulation environment

Yingjun Ye[a], Xiaohui Zhang[a], Jian Sun[a*]

[a] *Department of Traffic Engineering & Key Laboratory of Road and Traffic Engineering, Ministry of Education, Tongji University, Shanghai, China.*

**Abstract**

Automated vehicles (AVs) are deemed to be the key element for the intelligent transportation system in the future. Many studies have been made to improve the AVs' ability of environment recognition and vehicle control, while the attention paid to decision making is not enough though the decision algorithms so far are very preliminary. Therefore, a framework of the decision-making training and learning is put forward in this paper. It consists of two parts: the deep reinforcement learning (DRL) training program and the high-fidelity virtual simulation environment. Then the basic microscopic behavior, car-following (CF), is trained within this framework. In addition, theoretical analysis and experiments were conducted on setting reward function for accelerating training using DRL. The results show that on the premise of driving comfort, the efficiency of the trained AV increases 7.9% compared to the classical traffic model, intelligent driver model (IDM). Later on, on a more complex three-lane section, we trained the integrated model combines both CF and lane-changing (LC) behavior, the average speed further grows 2.4%. It indicates that our framework is effective for AV's decision-making learning.

## 1. Introduction

The automated vehicles have captured the public attention in recent years, especially after Google announced its automated driving program in 2010, for its advantages of alleviating the traffic congestion, liberating drivers' attention and conserving energy. The tasks involved in achieving autonomous driving can be divided into three modules: environment recognition, decision making and vehicle control. Among them, the vehicle control has no obvious differences between AV and manual driven vehicle. Meanwhile, the other two modules both turn to deep learning methods for better solutions in recent development. The AI-based environmental recognition methods including convolutional neutral network (CNN) (LeCun et al., 1998) and deep belief network (Hinton et al., 2006) are used to map from an image to perception information, such as detection of traffic participants (Szarvas et al., 2005; Wang et al., 2014), lane markings (McCall and Trivedi, 2004; Fritsch et al., 2013), traffic signs (De La Escalera et al., 1997; Sermanet and LeCun, 2011), etc. Study on end-to-end learning integrating three modules are also concerned in the autonomous driving field, that is to map raw pixels from a camera directly to steering commands (Bojarski et al., 2016). However, studies specified to decision-making task are very limited for various reasons. First, a large quantity of field data is the basis of training with deep learning algorithms, while data collection is a high-cost and time-consuming job. Many datasets, like KITTI (Geiger et al., 2013) and ImageNet (Deng et al., 2009), are used for training environment recognition, while no public data available for decision-making behaviors. Secondly, it is hard to label the decision-making behavior, given that actions are not only stimulated by the surrounding environment but also by the drivers' psychological factors (Van Winsum, 1999), while the latter cannot be easily observed and quantified. What's more, traditional imitation learning algorithms limit the operation skill to human understanding (Kober and Peter, 2010). That means, supervised learning from human driving data, can only imitate the driving behavior of sampled drivers, but not necessarily the optimal driving skill. And the generalization ability to a new environment is also doubtful.

[*] Corresponding author
  E-mail address: sunjian@tongji.edu.cn (J. Sun).

On the other hand, deep reinforcement learning (DRL) algorithm is proved to perform excellently on decision-making tasks. It solves the problems of relying on historical data and imitation to the human player, the resulting agent might act much better than a human does. Google's DeepMind team first successfully applied DRL to learn policies for agents on Atari 2600 games (Mnih et al., 2013; Mnih et al., 2015; Lillicrap et al., 2015; Mnih et al., 2016), with the results achieved even better performance than an expert human player. The final revolutionary achievement was the AlphaGo, an artificial intelligence computer program plays the Go, a board game that might be humankind's most complicated one. It indicated that DRL could achieve better performance than human beings in highly complex tasks.

It is known that three elements are critical to DRL: 1) Training environment. It should contain a variety of scenes to improve the generalization. And the rivals or interactive objects should be experts or good players at least, so that agent can learn from good examples. 2) Reinforcement learning algorithm. Each case corresponds to its own proper algorithm. For example, Q-learning is a simple but effective algorithm but it can be only adopted in the discrete-state problem. 3) Reward function. The training objective is incorporated into reward function to evaluate the reward or punishment of every action. However, the existing studies on autonomous driving have some insufficiencies in all these three aspects. The racing games (e.g. TORCS, GTA-V) are chosen as the training environment whose traffic flow scenes are not realistic. The input states and output actions are discretized, which seriously affects the accuracy of prediction. Furthermore, there are no uniform methods of reward function setting.

Therefore, based on high-fidelity simulation environment, we study the modeling of decision-making behavior with a modified DRL model in this paper. More specifically, there are mainly four contributions of this paper: (1) An integrated framework based on VISSIM high-fidelity simulation and DRL is proposed for decision-making training. (2) A DRL algorithm named deep deterministic policy gradient (DDPG) is adopted. This model is further improved to make it appropriate for autonomous driving with continuous state. (3) Three attempts of reward functions are investigated to explore their influence mechanism. What's more, the methods of verifying the effectiveness and two ways of accelerating convergence are discussed. (4) The proposed methods are first tested to train the CF model on a one-lane freeway section, and then extended to a three-lane section to learn the integrated model combines CF and LC behavior. Results indicate that it is effective to learn driving task with our platform.

The rest of paper is structured as follows. Section 2 reviews the recent work about the decision making of AV. Section 3 introduces the framework of our training platform and each component in detail. The case study of the CF behavior is trained in Section 4, and discussion is made on reward setting. An extended model integrates LC behavior is also presented. The last section presents the conclusion and future work.

## 2. Work related to AV's decision-making

The most influential competition of AVs is possibly the Grand Challenge launched by the Defense Advanced Research Projects Agency (DARPA). The first two competitions (2004, 2005) both conducted on one-way desert trails, thus more attention was paid on environment recognition and vehicle control to deal with different road conditions. Until these two modules were well-learned, the more complicated decisions making were involved in the third event, the Urban Challenge (2007). It called for AVs to drive through a much simplified urban environment, interacting with other vehicles and obeying the traffic rules. However, for getting higher rank, the competitors mostly used rule-based approaches (Buehler et al., 2009) to ensure absolute security, which was kind of conservative. Though great advances had been made compared to the first competition, the state of technology was still far from practical use.

It is generally known that simulation could offer simpler and less expensive alternatives, and allow direct comparison between agents under identical conditions. Therefore, after the DARPA Grand Challenge, to improve the decision-making behavior, international game AI competitions (WCCI-2008, CIG-2008, CEC-2009 and GECCO-2009) were held and drawn the public attention from field test to virtual simulator training. The Open Racing Car Simulator(TORCS) (Torcs, 2017) was one of the competition programs. Hand-crafted rule-based approaches were adopted in early research but resulted in limited performance (Buckland; Chen et al., 2015). To make it more efficient, researchers turned to artificial intelligence to find the solution. Driving behaviors were controlled with general machine learning approaches (Togelius, et al., 2006), like neural network and k-nearest neighbor, but still did not prove to give good results. Accordingly, The Neuro Evolution of Augmenting Topologies, tuning both the weighting parameters and structures of networks to find a balance between the fitness of evolved solutions and their diversity, had been applied to improve the performance of original neural network (Togelius et al., 2007; Cardamone et al., 2009; Cardamone et al., 2009). Other improvements of the machine learning approaches included enhancing the generalization capabilities of the evolved model (Kim et al., 2012), optimizing to multi-objectives (Van Hoorn, 2009), etc. These machine learning algorithms are able to learn driving to some extent, however, successful applications need large amounts of hand-labeled training data. Another issue is that those supervised learnings on human-driving data are limited to imitation to human drivers, which is able to drive but performed not necessarily excellent.

Until Google's DeepMind team introduced their DRL into behavior prediction, a totally new idea came into researchers' view. Since then, much similar work (Loiacono et al.,2010; Koutník et al., 2014; Lau ,2016; Sallab at al., 2017) had introduced DRL into TORCS driving. Meanwhile, some other games with more realistic graphics and other cars and pedestrians presented were also used to be the training platform, such as World Rally Championship 6 (WRC6) (Richter et al., 2016) and Grand Theft Auto V (GTA V) (Perot et al., 2017), but the results were still not satisfactory. They all focused on the single controlled vehicle and learned to keep in lanes merely, with few interactions with other traffic participants. The much more complex interaction with nearby vehicles on adjacent lane was rarely considered. Furthermore, the traffic scenes in those games only reproduce high-fidelity objects, while the characteristics of traffic flow were not really simulated.

## 3. Methodology

### 3.1 Framework of the training platform

The feasibility of training autonomous driving by DRL was verified in previous studies (Van Hoorn, 2009; Loiacono et al.,2010; Koutník et al., 2014; Lau ,2016; Richter et al., 2016; Perot et al., 2017; Sallab at al., 2017). However, the results were far from practical use for its simplifications from the driving environment to training target. Therefore, the training environment of previous studies, racing games, are replaced by traffic simulator VISSIM to provide more realistic interaction between traffic participants. And the DDPG algorithm is adopted which is suitable for the continuous decision spaces of both output states and input actions in autonomous driving.

As shown in Fig.1, the training platform of autonomous driving in this paper consists of two parts: the virtual simulation environment and the DRL training program. In every training step, the recognition information in VISSIM is passed to DRL program through the driving simulator interface. Based on the current environment, the DRL program evaluates the value of the last action and updates the network parameters. At the end of the step, the DRL program instructs the actions to the AV in VISSIM. The updates are done recursively with the steps going on until a feasible solution is found. Data is passed via TCP/IP communication protocol between the VISSIM and DRL program.

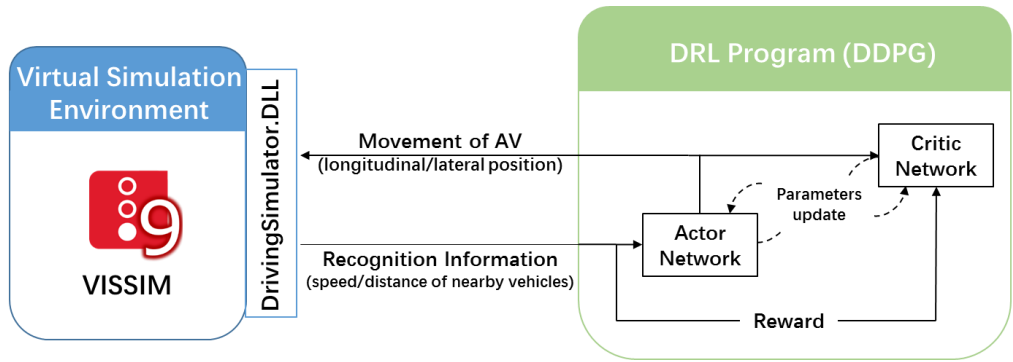

**Fig.1.** The framework of the training platform

*3.2 Training environment--VISSIM*

VISSIM is a widely used commercial traffic microscopic simulation software developed by the PTV (The developer of VISSIM). It provides various realistic traffic flow scenes including the road infrastructures and behavior of all road users, while the racing games focus on the 3D visual presentation (see Fig.2). With rich application programming interfaces, many studies have been made on the virtual testing of AVs in VISSIM.

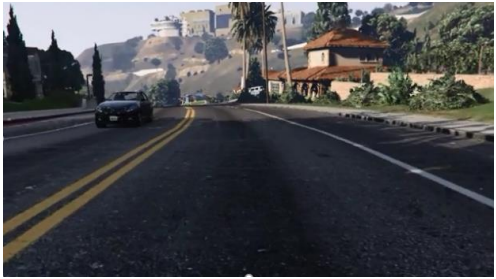 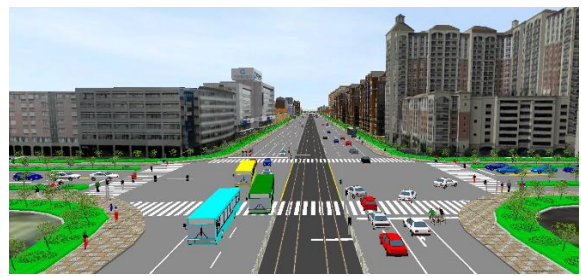

(a)  Screenshot of GTA V  (b)  Screenshot of VISSIM

**Fig.2.** Screen comparison between racing game and traffic simulator

*3.3 Training methods—DRL*

Reinforcement learning (RL) is defined as a sequential decision-making problem of an agent that has to learn how to perform a task through trial and error interactions with an unknown environment which provides feedback in terms of numerical reward (Van Hoorn, 2009). To solve the real problem, theoretical studies have continued to develop since the conception of RL is put forward. Therefore, the following parts firstly introduce the key approaches in the development of RL and their applications. Then the specific theoretical model in this paper is provided in detail.

*3.3.1 The development of RL*

**Markov Process** is the theoretical basis of the RL, which can be expressed as a tuple of {S, A, $P(a|s)$, γ, R}, where: S-set of states, A-set of actions, $P(a|s)$-state transition probability, namely the probability of transiting from state $s$ by taking action $a$, γ-discount factor, R-reward.

We use the mapping function $\pi(a|s) = P(a|s)$ to denote the policy that agent takes the action a when being in $a$ state $s$. Also we define a value function to define the expected total discounted rewards got by adopting this policy as Eq.(1), the objective is to find the optimal policy that maximizes the total reward.

$$Q_\pi(s, a) = E_\pi[\sum_{k=0}^{\infty} \gamma^k r_k | s, a] \tag{1}$$

***Q-learning*** (Watkins, 1989) is perhaps the most classic and commonly used algorithm to solve the MDP problem. It updates the action-value stored in Q-table based on temporal difference incremental learning, which can be expressed as:

$$Q(s_t, a_t) = Q(s_t, a_t) + \propto \left[ r(s_t, a_t) + \gamma \max_{a_{t+1}} Q(s_{t+1}, a_{t+1}) - Q(s_t, a_t) \right] \tag{2}$$

Where, $\propto$ is the learning rate. It can be only used to solve the MDP problems with actions and states space both discrete and finite. With an initial Q-table and predefined policy, in each step, the agent selects an action $a_t$ based on the current state $s_t$. According to the reward $r(s_t, a_t)$ feeds back and observation of the next state $s_{t+1}$, the Q-table is updated by Eq.(2). The steps repeat until Q function converge to an acceptable level.

Loiacono et. al. tried to learn to overtake through Q-learning (Loiacono et al., 2010). They designed two separated scenes: overtaking on a straight sketch and on a tight bend. But to reduce the action space, the agent only learns to steer wheels in the former scene while acceleration and brake are learned in the bend scene, thus the acceleration and steer are not integrated. Moreover, the discrete state and action space cannot precisely describe the interaction between vehicles.

***Deep Q Network (DQN).*** The aforementioned Q-learning cannot deal with the problem with continuous states and does not have generalization ability. For this reason, the Google DeepMind team put forward a DRL structure combines CNN and Q-learning (Mnih et al.,2013; Mnih et al., 2015). This method is applied on seven Atari 2600 games, with better performance on six of the games comparing to other previous approaches, and three of them surpass even a human expert.

In short, deep Q network replaces Q-table in Q-learning by the deep neural network (DNN), which compensate the limitation of finite states. The loss function of the DNN is defined as:

$$Loss = (r_s^a + \gamma \max_{a'} Q^\pi(s', a'; \theta^-) - Q^\pi(s, a; \theta))^2 \tag{3}$$

Where, $\theta^-$ and $\theta$ are parameters of target Q network and Q network. Target Q network is a lagging network, the final networks converge when the loss reaches a certain small value, that means the parameters of these two model is quiet close.

*3.3.2 Methodology of DDPG*

Deep Q learning achieves the input of the continuous state, but the output actions are still discrete, which does not meet the realistic demand. To solve this problem, the DeepMind team further proposed the ***DDPG*** (Lillicrap et al., 2015). It combines the DQN, the Deterministic Policy Gradient (DPG) and Actor-Critic algorithm.

As Fig.1 shows, the most revolutionary change is the introduction of the Actor-Critic algorithm. The Actor-Critic algorithm has two neural networks: 1) The critic network $Q(s, a; \omega)$ takes the action from actor network as the input, and outputs corresponding action value. It updates based on the loss of Q network, which is same as the Eq.(3). 2) The actor network uses the DPG to output action policy $P(a|s)$. To maximize the value of the current action, the parameters of actor network $\sigma$ update by gradient ascent, which uses the chain rule as follows:

$$\frac{\partial Q}{\partial \sigma} = \frac{\partial Q(s,a;\omega)}{\partial a} \times \frac{\partial a(s,\sigma)}{\partial \sigma} \tag{4}$$

With two networks performing their tasks independently meanwhile correlatively, not only the efficiency but also the accuracy is improved.

Besides, some other tricks such as experience replay are also incorporated to eliminate the data relevance. The basic idea is that by storing an agent's experiences, and then randomly drawing batches of them to train the network, we can more robustly learn to perform well in the task. By keeping the experience we draw random, we prevent the network from only learning about what it is immediately doing in the environment, and allow it to learn from a more varied array of past experiences. The Experience Replay buffer stores a fixed number of recent memories, and as new ones come in, old ones are removed. When the time comes to train, we simply draw a uniform batch of random memories from the buffer and train our network with them.

Based on DDPG, DeepMind team trained the virtual vehicle in TORCS and achieved excellent driving along the track (Lillicrap et al., 2015). Motivated by the successful demonstrations by DeepMind, some researchers did similar work and optimize the algorithm for handling partial observation and reducing computational complexity (Sallab et al., 2017), and also preliminary discussions are made about different reward functions (Lau, 2016). But these studies merely taught vehicle to drive along the track, the interaction with other traffic participants was not in consideration. What's more important, the reward function, which is actually critical to DRL's training and converge, are not provided with detailed setting methods, what we will discuss in next section.

## 4. Case Study

To testify the feasibility of our training platform and explore the tips about model training, firstly, we simplify the scene and select the CF behavior as the training object. Though CF behavior is deemed to be the most basic decision in driving, it is actually not easy to learn from scratch. With no driving experience and strict restrictions at all, the AV is expected to learn to follow the preceding vehicle with proper distance and steady acceleration. After successfully applied on CF behavior, our training platform is extended to more general driving tasks incorporating LC and overtaking behavior.

*4.1 Effectiveness and stability evaluation*

In supervised learning, one can easily track the performance of a model by evaluating it on the validation sets. In reinforcement learning, however, accurately evaluating the training effects can be challenging. Since the training objective is to maximize the total reward the agent collects averaged over a number of games. Thus the total reward or average Q value in one episode, and the loss of the critic network are commonly used in previous studies (Mnih et al.,2013; Lillicrap et al., 2015; Mnih et al.,2015) to evaluate the stability. Their reflections on the training convergence are essentially consistent. Nevertheless, these indexes fluctuate between episodes, that leads to a mathematical imprecise. Researchers only judge the convergence visually by line charts, though it is not objective enough. Therefore, we derive the approximated upper bound of average action value (refer to Appendix. A for the concrete process) to help to quantitatively evaluate the stability:

$$\lim_{t\to\infty} Q_t \approx \frac{r}{1-\gamma} \tag{4}$$

As for effectiveness, the score is the most intuitive method in previous game training. It is even compared to the score played by the human player to evaluate the intelligence of agent. The environment platform in this paper is a traffic simulator, with no scoring system. Meanwhile, there is no universal standard for autonomous driving evaluation. So, firstly, we evaluate the effectiveness of the trained driving strategy by visual observation. Then, for the model seems "effective", the IDM, a generally accepted CF model is used to be the evaluation standard.

*4.2 Variable definition*

Variables used in the experiments are summarized in Table 1.

**Table 1**

Symbols and parameters

| Variables and parameters | |
|---|---|
| $d$ | Net distance from preceding vehicle (m) |
| $v$ | Speed of the AV (km/h) |
| $\Delta v$ | Speed difference to preceding vehicle (km/h), $\Delta v = v - v_{pre}$, normalized by 40km/h |
| $\Delta a$ | Jerk (m/s$^2$) |
| $h_d$ | Desired time headway (s), $h_d$=2 |
| State flags | |
| C | Collision, *d*<0m |
| L | Low-speed following though preceding vehicle is far away, *d*>15m & *v*<1m/s & $\Delta v$<0m/s |
| U | Uncomfortable jerk, $\Delta a > 5.6 m/s^2$, it is the maximum jerk passenger cannot notice |
| R | Reverse, *v*<0m/s |
| O | Otherwise, vehicle moves in normal state |

*4.3 Training tips about model training*

It is known that some implementation details are the key factors of the deep learning (Wei, 2015), such as the initialization and fine-tune methods. However, regarding DRL, the major difference is the reward function, which has not been discussed in previous studies. Another problem of DRL is the consuming time, days of training is common especially for a complex issue.

Thus, some training tips about the reward setting and accelerated methods are necessary, which are discussed in this section.

The simulation environment in exploring cases are CF scenes with a 2 km-long single straight lane, with no signal control and other traffic facilities. When simulation begins, the AV comes into the road after 30s warm-up time. The simulation resolution is 0.1s. Three situations lead to a new simulation episode, those are the collision, unreasonable stop (far away from the front vehicle) or reaching the end of the road. Since we only study the decision-making module in our study, it is assumed that the environment information has been recognized by sensors, with no noise error. Recognition input is the current speed and acceleration of the AV, the spacing and speed difference to the preceding vehicle, movement information output is the longitudinal acceleration.

*4.3.1 Discussion on reward function*

The reward function represents the expected driving behavior. For example, it is known that for AV, the desired objectives are safety, high efficiency and driving comfort. Therefore, when the AVs are in unsafe state or drive unstably, the reward would be negative punishment. On the contrary, higher speed will receive a greater reward.

Even following the objectives above, the reward can still be unfeasible or inefficient if the specific function is improper. Therefore, three reward attempts are tested to find the most effective reward form. At first, two rewards with addition form and multiplication form are investigated respectively. Though the rewards seem rational, the results are not satisfactory. We find that it is caused by unmatched magnitude between reward and punishment. Thus, the normalized form and 100-time magnified form to normalized one are raised. But the latter cannot converge to a good result, which is finally solved by the regularization. The detailed discussion of each reward form is listed below.

a) Addition form and multiplication form

Our initial set of experiments use reward functions respectively with additional form $r_{a1}$ and multiplication form $r_{a2}$, which are listed as below:

$$r_{a1} = \begin{cases} -100 & C \text{ or } R \\ -80 & U \\ -50 & L \\ h_d \times v - d & O \end{cases} \quad (5)$$

$$r_{a2} = \begin{cases} -100 & C \text{ or } R \\ -80 & U \\ -50 & L \\ \dfrac{v}{|(d - h_d \times v) \times \Delta v|} & O \end{cases} \quad (6)$$

The punishments are totally same. They forbid collision(C) and reverse(R), the uncomfortable(U) or inefficient driving performances(L) are also panelized. Their rewards are different but similar to expecting higher speed and proper smaller distance. The reward in Eq.(6) is motivated by the constant time headway policies in ACC system, where vehicles are supposed to keep the speed close to the preceding vehicle and the distance proportional to speed. Though both of the reward functions seems to be rational, results are quite different.

According to the final performance, vehicles trained with the addition form $r_{a1}$ always accelerate to the desire speed until collide into the preceding vehicle. Vehicles with $r_{a2}$ are able to keep a proper distance, however, the frequent sharp speed-up and brake make the jerk to a uncomfortable state. Corresponding to the above results, as shown in Fig.3, the total rewards of $r_{a1}$ are always negative. Though rewards of $r_{a2}$ have reached a relative high positive value, it shows no sign of convergence.

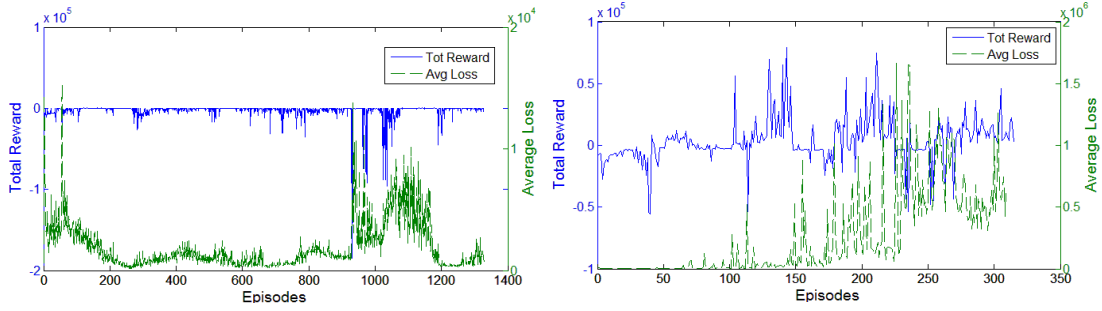

**Fig.3.** The total reward and average loss per episode with addition form on the left and multiplication form on the right

Even tried with different learning rates, no great progress has been presented. By theoretical analysis, we find the local optimal solution could be due to the magnitude gap between the reward value and punishment, which leads to a large step-size.

b) Normalized form and 100-time magnified normalized form

In the third reward function, we set the magnitude of the reward and punishment to the same level. Since it is difficult to define the boundary of the multiplication form, we normalize the reward of the addition form $r_{a1}$, the other variant is magnifying the normalized form by 100:

$$r_{a3} = \begin{cases} -1 & C\ or\ R \\ -0.8 & U \\ -0.5 & L \\ v' - d' & O \end{cases} \qquad (7)$$

$$r_{a4} = \begin{cases} -100 & C\ or\ R \\ -80 & U \\ -50 & L \\ 100 * (v' - d') & O \end{cases} \qquad (8)$$

Where, $d'$ and $v'$ are normalized variables of $d$ and $v$, respectively normalized by 100m and 80km/h.

Training with $r_{a3}$, the learning procedure converges after about 800 episodes(see in Fig.4). What's more, the resulting automated vehicles are able to drive with safety and comfort. In other words, vehicles could keep a proper distance from preceding vehicle, meanwhile avoid frequent or sharp acceleration and brake. Besides, assuming that the expected headway is 2s, according to the upper bound of action value evaluated by Eq.(4):

$$max(r_{a3}) \approx \frac{80}{80} - \frac{2 * 80/3.6}{100} = 0.56 \qquad (8)$$

$$Q_{a3}^{up} = \lim_{t\to\infty} Q_t \approx \frac{max(r_{a3})}{1-\gamma} = \frac{0.56}{0.01} = 56 \qquad (9)$$

The maximum of average Q value in our training procedure is 42.0 which is close to the evaluated upper bound. It verifies that normalized form is an effective reward function.

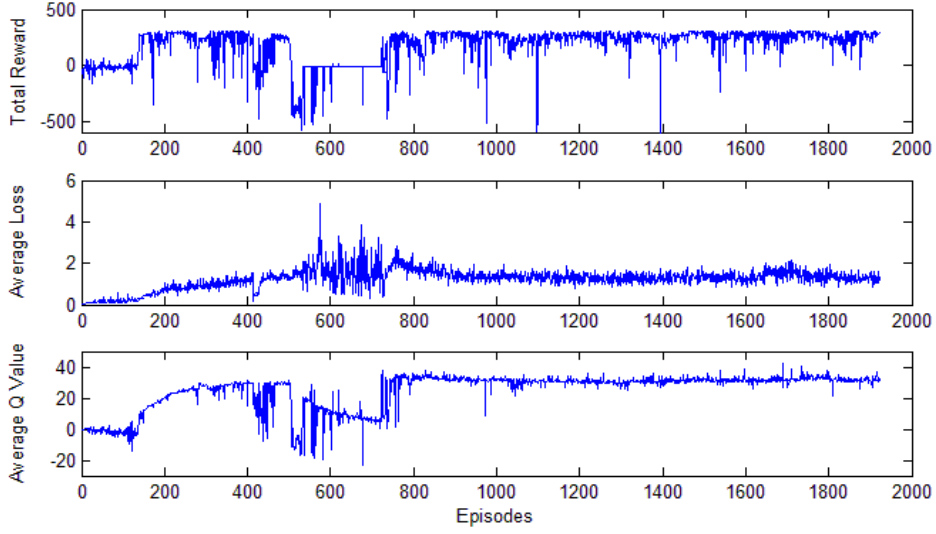

**Fig.4**. The evaluation of the normalized reward

However, the 100-time magnified form $r_{a4}$ almost learn nothing, the action value even converge to a negative value.

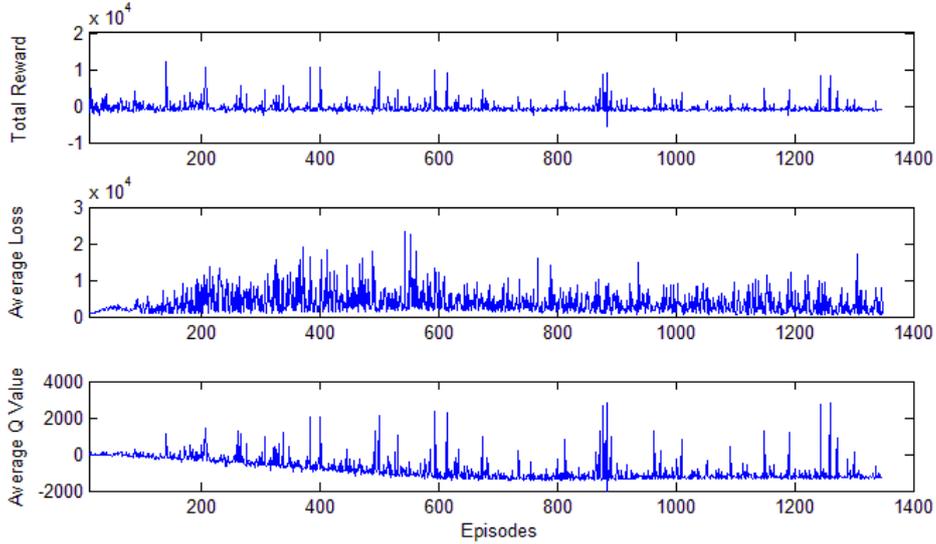

**Fig.5.** The evaluation of the 100-time magnified normalized form

c) Regularized bounded form

Further analysis of the training mechanism, the reason that 100-time magnified normalized form almost learns nothing is caused by the common problem of ReLU, the activation function of critic network. After 100-time magnification, the previous learning rate is so big that "dying ReLU" problem is easily raised and then results in a local solution. According to the parameter $\omega$ update

algorithm of critic network Q:

$$\omega \leftarrow \omega - \alpha \times \frac{\partial Loss}{\partial \omega} \qquad (10)$$

$$\frac{\partial Loss}{\partial \omega} = -2 \times \sqrt{Loss} \times \frac{\partial Q^\pi(s,a;\omega)}{\partial \omega} \qquad (11)$$

Where, $\alpha$ is learning rate. It is clear that the step-size of update parameter is proportional to square root of loss. The magnification of reward also leads to magnification of loss. Hence, to keep the update step-size to a proper level, the learning rate should be limited to a relatively low level.

On the other hand, the magnification also affects the action network. The activation function of action network, *tanh* function might also fall into local solution when the step-size is too big. Regularize the action network is effective for this problem, that is introducing parameters of action network into loss.

To validate our theory, we still adopt the reward $r_{a4}$, but minify the learning rate to 1% of original value. At the same time, regularize the action network. The result is similar to the normalized form. The model converges about 700 episodes' training.

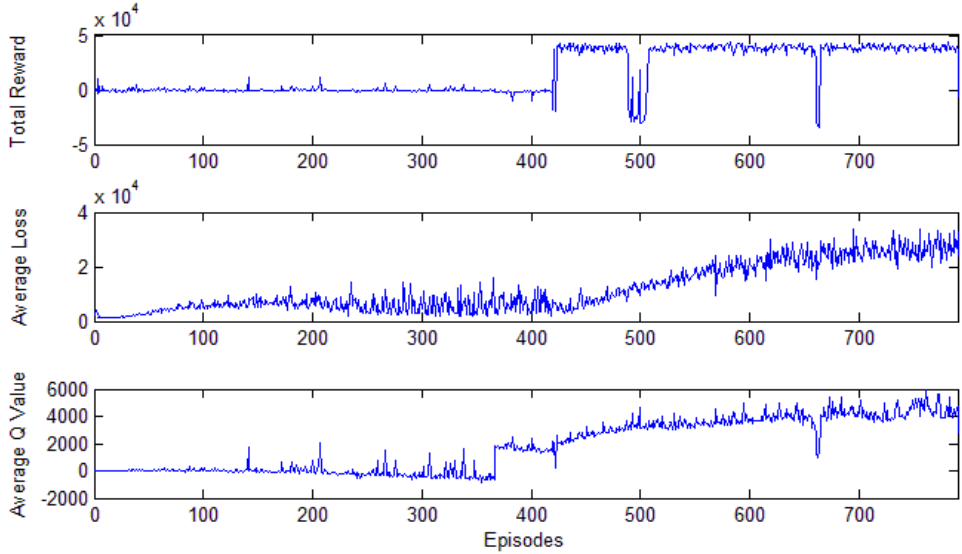

**Fig.6.** The evaluation of the regularized bounded form

Therefore, to make the model effective and quickly converge, there are mainly three key points on setting the reward function: 1. keep the reward and punishment bounded; 2. adjust the learning rate; 3. regularize the action network.

*4.3.2 Ways to accelerate training*

Generally, due to the high computational complexity, it spends considerable time on training the model. What' more important, the optimization of network parameters takes dozens of times' training. Apart from the hardware acceleration, we introduce two accelerated tips by primary trained parameter and enlarged time step.

a)  Based on primary trained parameter

The previous training all "learn from zero", that is automated vehicle has no any driving experience which leads to frequent collision or long following headway. It's easy to understand that it takes a rather long time from beginner to skilled "driver". So, we consider whether it is possible to "learn from the experienced" and then get further improvement. For this purpose, firstly we train the action network with the same dataset of IDM's calibration. As shown in Fig.7, it converges after about 400 episodes, while it takes almost 800 episodes to learn from zero. Therefore, the DRL based on primary trained parameter do accelerate training.

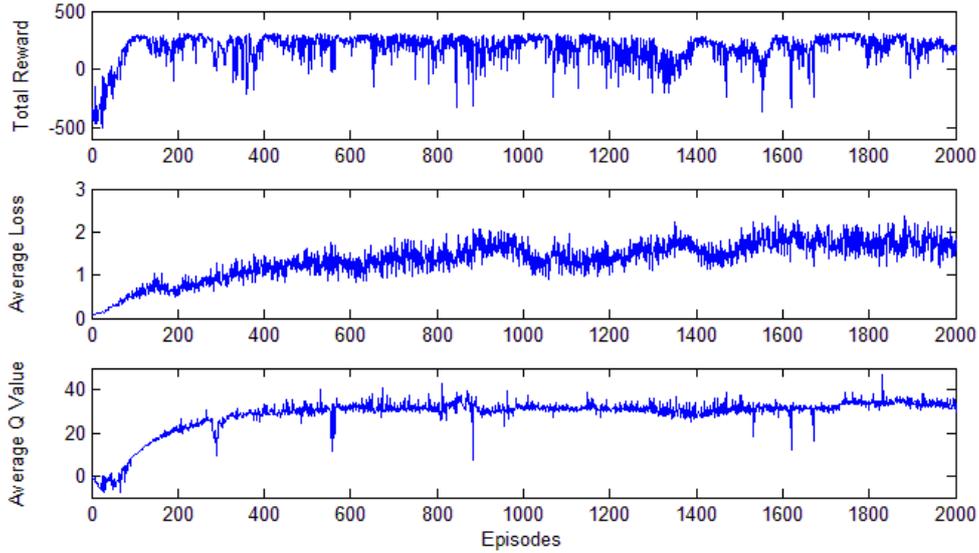

**Fig.7.** The evaluation of the normalized form with pre-trained parameter

b)  Properly enlarge time step

All the time steps in previous cases are set as 0.1s, that is to say, all the vehicles in the network update their status for every 0.1s. Considering human drivers do not adjust their actions every moment, but continue for some time after every action instead. So, by properly enlarging time step, we can shorten the training time in the premise of assuring effectiveness. Theoretically, if we enlarge the time step to 1s, it leads to 10-fold increases in efficiency. Results in the Fig.8 show this method is feasible, the 1s model converges at about 800 episodes which are similar to the 0.1s model.

*4.4 Evaluation of the DRL trained CF model*

On the previous discussion about reward function, we have tentatively verified the effectiveness of the DRL model. To quantify the evaluation of this model, we compare the DRL to IDM, a widely used CF model, about the efficiency and comfort. The DRL model is trained with the normalized reward $r_{a3}$.

Regarding efficiency, we extract the 800 episodes of DRL and IDM respectively and assess their average speed for each episode. For DRL model, we take the last 800 converged episodes of normalized reward. The IDM is calibrated with the microscopic trajectories collected from

VISSIM whose environment is totally same as the DRL training. As we can see from Fig.9, that the behavior of DRL is relatively decentralized but more aggressive. The average speed of DRL (12.44m/s) is 7.9% higher than IDM(11.53m/s).

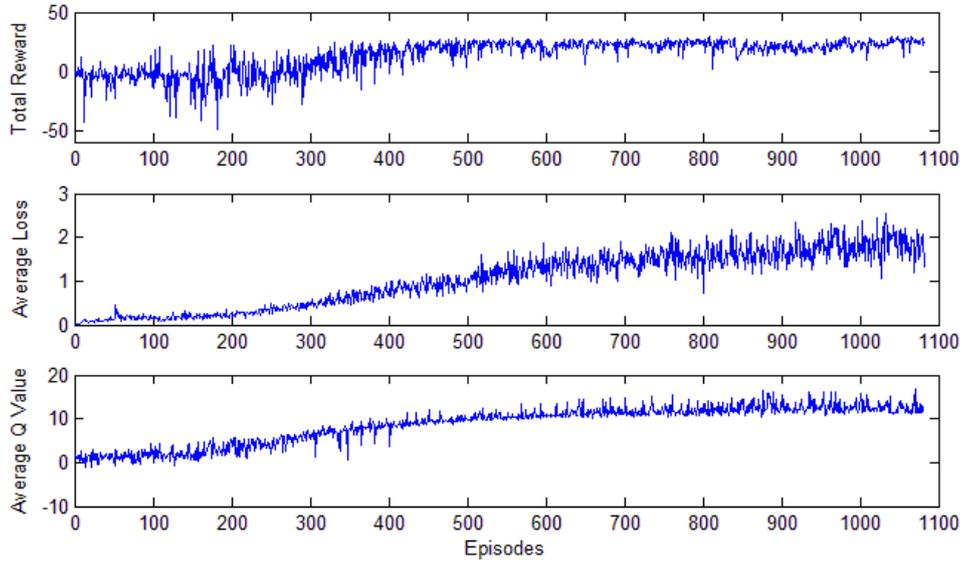

**Fig.8.** The evaluation of the normalized form with big time step

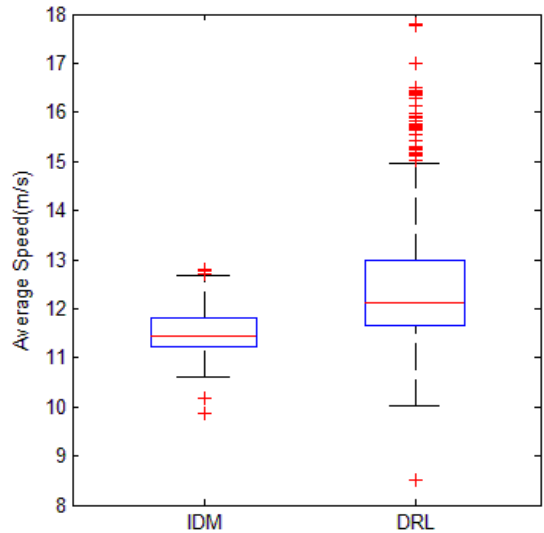

**Fig.9.** The average speed distribution of the IDM and DRL CF model

Comfort is assessed by jerk in this study. To figure it more clearly, we further extract 2000 steps' jerk randomly from the 800 episodes. As shown in Fig.10, though it is obvious that the jerks of DRL are much higher than IDM, the general distribution can meet the demand of human ride comfort (no higher than 5.6 m/s$^2$, dashed green line). Only 2.6% of the jerks exceed the critical value.

Therefore, we can conclude that on the premise of comfortable ride, the DRL model is more effective than IDM. The whole framework of the decision-making training is feasible.

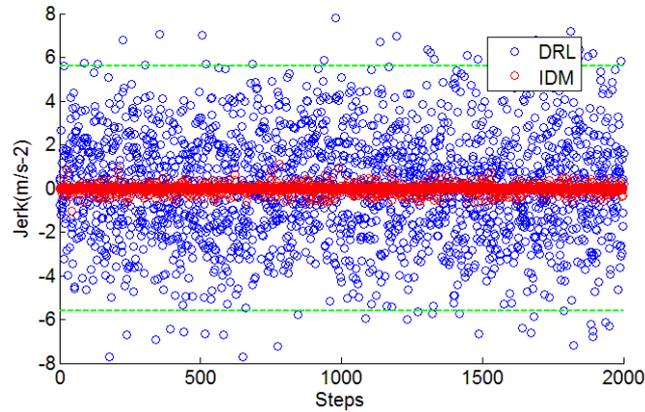

**Fig.10.** The jerk distribution of the IDM and DRL CF model

*4.5 Extended model integrating CF and LC behavior*

Compared to CF, LC behavior is a more challenging driving task, for considering more interactive objects on the road. To testify our platform on a more complex environment, and train a model capable of handling most of the traffic scene, we extended the model integrate CF and LC behavior on a three-lane section. The flow input is set as 700veh/h/ln, which is neither heavy nor low for more lane-changing intention. There are two major differences in comparison to CF training:

**Input and output of the model.** Since LC behavior focuses not only vehicles on the current lane, but also interactive objects on adjacent lanes. Thus, environment recognition input contains the current speed and acceleration of the AV, the lateral and longitudinal spacing and longitudinal speed information of each nearby vehicle (as shown in Fig.11, subject vehicle is the blue one, nearby vehicles one downstream and one upstream, 2 lanes on both sides and the current lane are marked). As for movement information output, besides longitudinal acceleration, LC instruction is also provided (3 discrete states: lane-keeping (LK), left lane-changing (LLC), and right lane-changing (RLC)). With a random output of sigmoid activation function, the initial probability of three decisions on LC is identical.

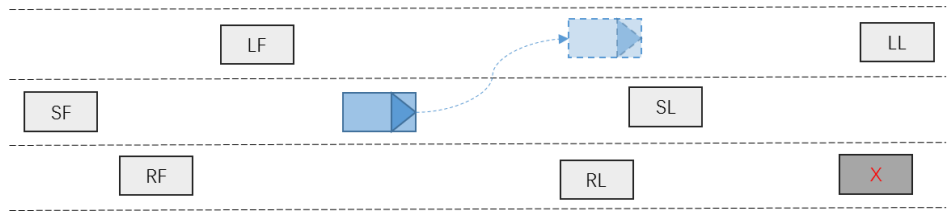

**Fig.11.** Description of the nearby vehicles of a LC process

**Reward function.** LC reward is also incorporated referring to MOBIL model (Kesting et al., 2007), which involves speed of the AV $v'$, the acceleration increment of the AV and followers on current and subject lane (e.g., if AV in Fig.11 tends to make left lane-change, the followers are SF and LF). That means, if LC behavior leads sharp brake to followers, even though the AV is accelerated, it will not get a great reward, and even be punished. What's more, when the gap is too small for the AV to cut-in, there will be a punishment (G) and AV will not change the lane. Multiple functions have been tried, the best performed one is listed as below:

$$r = \begin{cases} -1 & C \text{ or } R \\ -0.8 & U \text{ or } G \\ -0.5 & L \\ v' & car-following\ reward \\ 2*v' - 0.1*(\Delta a + \Delta a_{CF} + \Delta a_{LF}) & lane-changing\ reward \end{cases} \quad (12)$$

In order to simplify the model, on the process of AV's lane-changing, we assume that AV follows the nearer vehicle on both current lane and the target lane for the entire LC process (set to be 2s in this paper).

*4.5.1 Model stability*

As shown in Fig.12, the evaluations on reward, loss and Q value converge over 5000 episodes. It is necessary to explain that the slight fluctuations after convergence are acceptable, which could also be found on other studies (Mnih et al., 2013; Mnih et al., 2015).

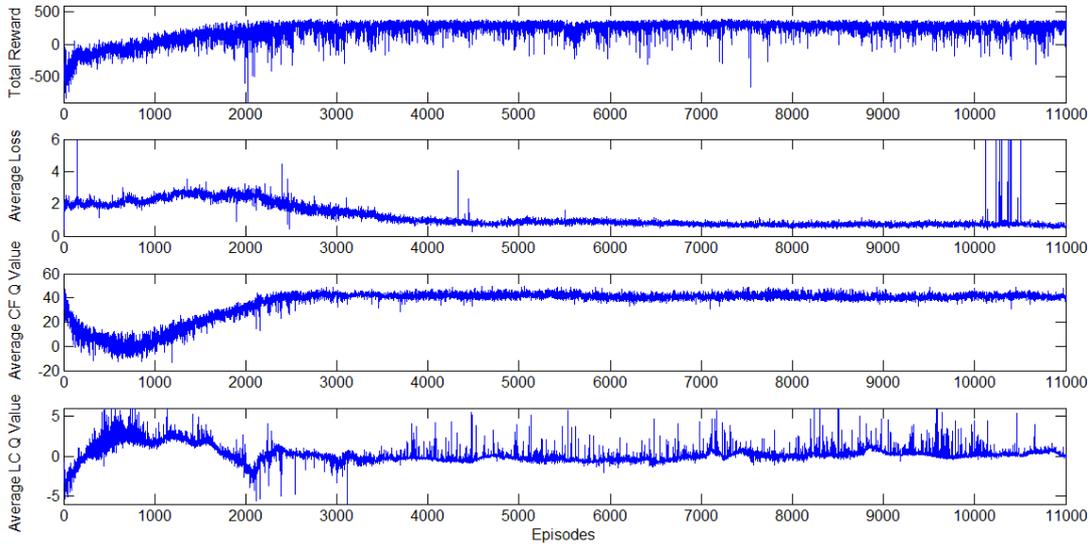

**Fig.12.** The average speed distribution of the integrated model

*4.5.2 LC decision analysis*

To better understand the improvement of lane-changing decision along the training process, the LC success rate (failed LC happens when the gap is too small even if the LC instruction is given, corresponds to state G) and average reward of successful LC are introduced. We can see from the Table 2, as the number of episodes increases, the frequency of LC instruction within each episode decreases from 267 to 20. The LC success rate, which means the ratio of successful LC decision among all the LC instruction, upgrades firstly and then descends to a stable state around 4%. It indicates that at the beginning, the AV is instructed with LC order frequently (2/3 of the instructions are LC, for the random ratio among LK, LLC and RLC). After thousand episodes of training, it learns not to LC at some wrong moment. What's more, the increase of average reward of success LC, and decrease of actual LC frequency in simulation both further prove the optimization of lane-changing decision. According to our statistics, the LC frequencies of two

NGSIM datasets are 0.63 veh/km/ln (US-101) and 0.80 veh/km/ln (I-80), which are similar to our converged result, 0.81 veh/km/ln. However, the traffic flow is heavier in the NGSIM data (around 1600 veh/ln on US-101 and 1300 veh/ln on I-80) comparing our simulation (around 700 veh/ln). Actually, higher LC frequency is desirable since we observe that in some scenarios, the AV is expected to make LC but still follow the front vehicle.

**Table 2**

Evaluation of LC training

| Episode | Avg. Freq. of LC instruction | LC success rate (success LC/LC instruction) | Avg. reward of success LC | Actual LC Freq. (veh/km/ln) | |
|---|---|---|---|---|---|
| <=50 | 267 | 3.2% | 2.29 | 8.62 | |
| <=100 | 266 | 3.3% | 2.63 | 8.72 | 9.26 |
| <=500 | 235 | 4.2% | 2.86 | 9.77 | |
| 500−1000 | 127 | 7.5% | 3.27 | 9.43 | |
| 1000−2000 | 51 | 8.0% | 3.60 | 4.06 | 4.06 |
| 2000−3000 | 51 | 3.5% | 4.92 | 1.82 | 1.82 |
| 3000−4000 | 27 | 4.4% | 5.29 | 1.18 | 1.18 |
| 4000−5000 | 21 | 5.6% | 5.76 | 1.19 | |
| 5000−6000 | 22 | 4.7% | 5.74 | 1.02 | |
| 6000−7000 | 15 | 5.9% | 6.91 | 0.89 | |
| 7000−8000 | 17 | 4.5% | 6.80 | 0.77 | 0.81 |
| 8000−9000 | 13 | 5.8% | 6.78 | 0.73 | |
| 9000−10000 | 17 | 4.0% | 5.73 | 0.66 | |
| 10000−11000 | 16 | 4.3% | 6.50 | 0.69 | |

The lead/lag time headways of the LC moment are presented in Table 3. Both the lead time headway and lag headway are significantly higher than the field data of NGSIM dataset. It might result from our LC reward, which considering not only the speed increase of the AV but also the lag vehicles. The objective of our LC model is improving the overall efficiency, while human driver is more willing to get a better driving condition on his own.

**Table 3**

Evaluation of LC gap

| | Avg. lead time headway (s) | Avg. lag time headway (s) |
|---|---|---|
| US-101 | 2.78 | 3.14 |
| I-80 | 1.49 | 1.71 |
| DRL-LC | 4.70 | 4.81 |

*4.5.3 Model effectiveness*

The average speed of this integrated model (12.74m/s) is even 2.4% higher than that of the DRL CF model. Therefore, the extended model integrating CF and LC behavior provides a more efficient solution.

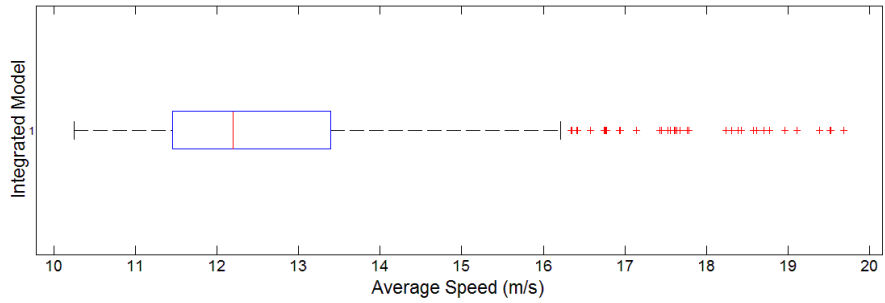

**Fig.13.** The average speed distribution of the integrated model

In general, the integrated model can provide more efficient driving decision, but the LC behavior so far is still kind of conservative. Therefore, modification on LC reward and thousands of training episodes are required to refine a more rational LC decision, and more specific analysis should be made in our later study.

## 5. Conclusion

For AVs, subject to the insufficient available data and limitation of the conventional training method, learning an efficient and smart decision-making algorithm seems unachievable. Therefore, we proposed a framework for decision-making learning in this paper, which solve the problems above by introducing the DRL and high-fidelity virtual simulation environment. DDPG and VISSIM are adopted for each of their parts. Firstly, to ensure the convergence, the key element of DRL reward function, are explored for various forms. It concludes that the bounded reward, learning rate and regularized action network should be noted on setting reward. For training acceleration, the primary parameter trained by deep learning and enlarged time step are available. Then, in the aspect of comfort and effectiveness, the trained DRL model is evaluated compared to IDM. Results show that even though it is more comfortable with the IDM, only 2.6% of the jerks exceed the comfort threshold, thus the overall process of DRL model is acceptable. As for efficiency, the average speed of DRL is even 7.9% lower than the IDM's. Extending to the more complex task, training a model integrates LC and CF behavior, vehicles perform even better efficiency. The average speed gets 2.4% increase compared to the solely DRL-trained CF model. In consequence, the training framework proposed in this paper for decision making is effective. And training the more robust integrated model will be our future work.

**Acknowledgments**

The authors would like to thank the Natural Science Foundation of China (51422812), and the Shanghai Science and technology project of international cooperation (16510711400) for supporting this research. We also would like to thanks PTV to supply the VISSIM Engine for accelerating our DRL training program.

**Appendix A. Estimation of the approximated upper bound of average action value**

**Proposition 1.** When evaluating the convergence of the current prediction, an upper bound is referred to avoid instability of the commonly used three indexes. The upper bound is expressed as:

$$\lim_{t\to\infty} Q_t \approx \frac{r}{1-\gamma} \quad (A1)$$

**Proof.** According to Eq.(1), we expand the action value as:

$$Q^\pi(s,a) = E(r_t + \gamma r_{t+1} + \gamma^2 r_{t+2} + \cdots | s,a)$$
$$= E(r_s^a + \gamma Q^\pi(s',a') | s,a) \quad (A2)$$
$$\approx r_s^a + \gamma Q^\pi(s',a')$$

Denote

$$Q_t = Q^\pi(s,a) \quad (A3)$$
$$Q_{t+1} = Q^\pi(s',a') \quad (A4)$$
$$r_t = r_s^a \quad (A5)$$
$$r = max(r_t) \quad (A6)$$

So, we have

$$Q_t \approx r_t + \gamma Q_{t+1} \quad (A7)$$

When the training converges, every step is deemed to have maximized reward. Thus for any step $t$,

$$Q_t \approx r + \gamma Q_{t+1} \quad (A8)$$
$$Q_t \approx Q_{t+1}$$

Then, we have the approximated upper bound of average action value:

$$\lim_{t\to\infty} Q_t \approx \frac{r}{1-\gamma} \quad (A9)$$